\documentclass{svproc}
\usepackage{url}
\usepackage{amsmath}
\usepackage{siunitx,booktabs}
\usepackage{epsfig}
\usepackage{algorithm}
\usepackage{algorithmic}
\usepackage{multirow}

\begin{document}
\mainmatter              
\title{Applying ACO To Large Scale TSP Instances}
\titlerunning{Applying ACO To Large Scale TSP Instances}  
%
\author{Darren M. Chitty}
\authorrunning{Darren M. Chitty} 
%
\tocauthor{Darren M. Chitty}

\institute{Department of Computer Science,\\ University of Bristol, Merchant Venturers Bldg,\\ Woodland Road, BRISTOL BS8 1UB\\ \email{darrenchitty@googlemail.com}}  \maketitle              

\begin{abstract}
Ant Colony Optimisation (ACO) is a well known metaheuristic that
has proven successful at solving Travelling Salesman Problems
(TSP). However, ACO suffers from two issues; the first is that the
technique has significant memory requirements for storing
pheromone levels on edges between cities and second, the iterative
probabilistic nature of choosing which city to visit next at every
step is computationally expensive. This restricts ACO from solving
larger TSP instances. This paper will present a methodology for
deploying ACO on larger TSP instances by removing the high memory
requirements, exploiting parallel CPU hardware and introducing a
significant efficiency saving measure. The approach results in
greater accuracy and speed. This enables the proposed ACO approach
to tackle TSP instances of up to 200K cities within reasonable
timescales using a single CPU. Speedups of as much as 1200 fold
are achieved by the technique.

\end{abstract}

\section{Introduction}
Ant Colony Optimisation (ACO) \cite{dorigo:2004} is a
metaheuristic which has demonstrated significant success in
solving Travelling Salesman Problems (TSP) \cite{dorigo:1997}. The
technique simulates ants moving through a fully connected network
using pheromone levels to guide their choices of which cities to
visit next to build a complete tour. However, ACO has two
drawbacks the first being significant memory requirements to store
the pheromone levels on every edge. Secondly, simulating ants by
making probabilistic decisions at each city to determine the next
city to visit makes ACO computationally intensive. Therefore, ACO
will struggle when applied to larger TSP instances. Consider, the
pheromone matrix which requires an $n$ by $n$ matrix whereby $n$
is the number of cities. As the number of cities increases
linearly, a quadratic increase in memory requirements is observed.
The same is true for probabilistically simulating ants to
construct a tour. This paper will address these issues enabling
ACO to be applied to larger TSP instances.

The paper is laid out as follows; Section \ref{sec:ACO} will
describe ACO, Section \ref{sec:scale} will present a scalable
version of ACO to apply to large scale TSP instances whilst
Section \ref{sec:Experiments} will demonstrate its effectiveness
on well known TSP instances. Finally Section \ref{sec:LargeTSP}
demonstrates the approach on TSP instances of up to 200,000
cities.

\section{ACO Applied to the TSP} \label{sec:ACO}
The Travelling Salesman Problem (TSP) is a task where the
objective is to visit every city in the problem once minimising
the total distance travelled. The symmetric TSP can be represented
as a complete weighted graph $G=(V,E,d)$ where $V=\{1,2,..,n\}$ is
a set of vertices defining each city and $E=\{(i, j)|(i, j)\in V
\times V\}$ the edges consisting of the distance $d$ between pairs
of cities such that $d_{ij}=d_{ji}$. The objective is to find a
Hamiltonian cycle in $G$ of minimal length.

Ant Colony Optimisation (ACO) applied to the TSP involves
simulated ants moving through the graph $G$ visiting each city
once and depositing pheromone as they go. The level of pheromone
deposited is defined by the quality of the tour the given ant
finds. Ants probabilistically decide which city to visit next
using this pheromone level on the edges of graph $G$ and heuristic
information based upon the distance between an ant's current city
and unvisited cities. An \emph{evaporation} effect is used to
prevent pheromone levels reaching a state of local optima.
Therefore, ACO consists of two stages, the first \emph{tour
construction} and the second stage \emph{pheromone update}. The
tour construction stage involves $m$ ants constructing complete
tours. Ants start at a random city and iteratively make
probabilistic choices as to which city to visit next using the
\emph{random proportional rule} whereby the probability of ant $k$
at city $i$ visiting city $j\in N^{k}$ is defined as:
\vspace{-0.75cm}

\begin{eqnarray}
    p_{ij}^{k}=\frac{[\tau_{ij}]^{\alpha}[\eta_{ij}]^{\beta}}{\sum_{l\in
N^{k}}[\tau_{il}]^{\alpha}[\eta_{il}]^{\beta}}
\end{eqnarray}

where $[\tau_{il}]$ is the pheromone level deposited on the edge
leading from city $i$ to city $l$; $[\eta_{il}]$ is the heuristic
information consisting of the distance between city $i$ and city
$l$ set at $1/d_{il}$; $\alpha$ and $\beta$ are tuning parameters
controlling the relative influence of the pheromone deposit $
[\tau_{il}]$ and the heuristic information $[\eta_{il}]$.

Once all ants have completed the tour construction stage,
pheromone levels on the edges of graph $G$ are updated. First,
evaporation of pheromone levels upon every edge of graph $G$
occurs whereby the level is reduced by a value $\rho$ relative to
the pheromone upon that edge:
\begin{eqnarray}
\tau_{ij}\leftarrow(1-\rho)\tau_{ij}
\end{eqnarray}

where $p$ is the \emph{evaporation rate} typically set between 0
and 1. Once this evaporation is completed each ant $k$ will then
deposit pheromone on the edges it has traversed based on the
quality of the tour it found:
\begin{eqnarray}
\tau_{ij}\leftarrow\tau_{ij}+\sum_{k=1}^{m}\Delta \tau_{ij}^{k}
\end{eqnarray}

where the amount of pheromone ant k deposits, $\Delta
\tau_{ij}^{k}$ is defined by:
\begin{eqnarray}
\Delta \tau_{ij}^{k}&=&\left\{
\begin{array}{ll}
1/C^{k}, & \mbox{if edge $(i,j)$ belongs to $T^{k}$} \\
0, & \mbox{otherwise} \\
\end{array}
\right.
\end{eqnarray}

where $1/C^{k}$ is the length of ant $k$'s tour $T^{k}$. This
methodology ensures that shorter tours found by an ant result in
greater levels of pheromone being deposited on the edge of the
given tour.

\section{Addressing the Scalability of ACO} \label{sec:scale}
A key issue with ACO is its memory requirements in the form of the
pheromone matrix which stores the the level of pheromone on every
edge between each city. Thus an $n$ by $n$ size matrix is required
in memory to store this information so for a 100,000 city problem,
a 100,000 by 100,000 matrix is required. Using a float datatype
requiring four bytes of memory, this matrix will need
approximately 37 GB of memory, much more than typically available
on CPUs. However, a variant of ACO exists which dispenses with the
need for a pheromone matrix, Population-based ACO (P-ACO)
\cite{guntsch:2002}. With this approach, a population of tours are
maintained ($k_{long}(t)$) whereby the best tour at each iteration
$t$ is added. Since $k_{long}(t)$ is of a fixed size tours are
added in a First In First Out (FIFO) manner. Pheromone levels are
calculated by using the $k_{long}(t)$ information. An ant at a
given city calculates the pheromone levels by examining the edges
that were traversed in $k_{long}(t)$ from the given city. Thus
there is no pheromone matrix and no pheromone evaporation. If
$k_{long}(t)$ is significantly less than the number of cities then
this is a considerable saving in memory requirements.

In this paper, some modifications to P-ACO are implemented.
Firstly, instead of using a store of best found tours updated in a
FIFO manner, each ant has a \emph{local memory} ($l_{best}$)
containing the best tour that the ant has found, a
\emph{steady-state} mechanism. These $l_{best}$ tours are used to
provide pheromone level information to ants when probabilistically
deciding which city to next visit. This is similar in effect to
Particle Swarm Optimisation (PSO) \cite{eberhart:1995} whereby
particles use both their local best solution and a global best to
update their position. Secondly, the amount of pheromone an edge
from an $l_{best}$ tour contributes equates to the quality of the
\emph{global best} ($g_{best}$) tour divided by that of the
$l_{best}$ hence a value between 0.0 and 1.0. These measures are
taken to increase diversity.

Moreover, to gain the maximum available performance of the P-ACO
approach from a CPU, an asynchronous parallel approach is used
with multiple threads of execution and each thread simulates a
number of ants. Moreover, the choosing of the next city to visit
is decided by multiplying the heuristic information, the pheromone
level and a random probabilistic value between 0.0 and 1.0 and the
city with the greatest combined value is selected as the next to
visit. This approach is known as the \emph{Independent Roulette}
approach \cite{cecilia:2013}. This allows the utilisation of the
extended Single Instruction Multiple Data (SIMD) registers
available in a CPU through AVX for the probabilistic decision
making process. These extra wide registers enable up to eight edge
comparisons to be made in parallel. Using a parallel methodology
with AVX registers improves the computational speed by
approximately 30-40x when using a quad core processor.

\subsection{Introducing \emph{PartialACO}}
\begin{figure}
\centering
{\setlength{\fboxrule}{1pt}\fbox{\includegraphics[width=0.35\textwidth]{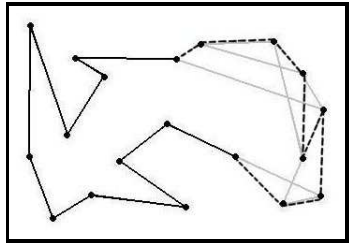}}}
\footnotesize \caption{Illustration of \emph{PartialACO} whereby
the $l_{best}$ tour of an ant is partially modified by the ant.
The dark part of the tour is retained and the lighter part
discarded. \emph{PartialACO} then completes the tour as normal
creating a new tour (dashed line).} \label{fig:PartialACO}
\end{figure}

The most computationally expensive aspect of the ACO algorithm is
the \emph{tour construction} phase. This aspect of ACO has an
exponential increase in computation time cost as the number of
cities increases. Moreover, as an ant repeatedly probabilistically
decides at each city which to visit next it could be considered
that the greater number of cities in a tour that requires
constructing, the greater the probability an ant will eventually
make a poor choice of city to visit next resulting in a low
quality tour. Hence, it is hypothesised that perhaps it would be
advantageous for ants to only change part of a known good tour. To
do so would firstly reduce the computational complexity and
secondly reduce the probability of an ant making a poor decision
at some point of the tour construction. For the P-ACO approach
detailed previously, the part of the tour that is not changed by
an ant could be based upon its $l_{best}$ tour. Essentially, at
each iteration an ant randomly chooses a city to start its tour
from and then a random number of cities to preserve from its
$l_{best}$ tour. The remaining part of the tour will be
constructed as normal. This methodology is similar to crossover in
Genetic Algorithms (GA) \cite{Holland:1975} for the TSP whereby a
large section of a tour is preserved and the remaining aspect
constructed from another tour whilst avoiding repetition. Figure
\ref{fig:PartialACO} visualises the concept whereby the dark part
of the $l_{best}$ tour is preserved and the rest discarded and
then this partial tour is completed using ACO. Henceforth, this
implementation of ACO will be referred to as \emph{PartialACO}. A
high level overview of the technique is shown in Algorithm
\ref{alg:PartialACO}.

\vspace{-0.5cm}

\vspace{-1.0cm}

\begin{algorithm}[]
\caption{\emph{PartialACO}} \label{alg:PartialACO}
\begin{algorithmic}[1]
    \FOR{each ant}
        \STATE{Generate initial tour using P-ACO approach}
    \ENDFOR
    \FOR{number of iterations}
        \FOR{each ant}
            \STATE{Select random starting city from current $l_{best}$ tour}
        \STATE{Select random number of cities in $l_{best}$ tour to preserve}
        \STATE{Copy $l_{best}$ tour from starting city for the specified random number of cities}
        \STATE{Complete remaining aspect of tour using P-ACO approach}
        \STATE{If new tour better than $l_{best}$ tour then update $l_{best}$ tour}
    \ENDFOR
\ENDFOR \STATE{Output best $l_{best}$ tour (the $g_{best}$ tour)}
\end{algorithmic}
\end{algorithm}

\section{Experiments With \emph{PartialACO}}
\label{sec:Experiments} To test the effectiveness of the
\emph{PartialACO} approach experiments will be conducted using
five standard TSP problems of increasing size from the TSPLIB
library. Sixteen ants, two per parallel thread of execution, will
be simulated for 100,000 iterations with the $\alpha$ and $\beta$
parameters both set to a value of 5.0 to reduce the influence of
heuristic information and increase the influence of pheromone from
good tours. Results are averaged over 100 random runs and
experiments are conducted using an Intel i7 processor using eight
parallel threads of execution and the AVX registers. Table
\ref{tab:ACO} shows the results from the standard P-ACO approach
whereby full length tours are constructed by each ant at every
iteration. The average accuracy ranges from between 4 and 13\% of
the known optimum. Table \ref{tab:cross} demonstrates the results
from the \emph{PartialACO} approach described in this paper
whereby only a portion of each ant's best found tour is exposed to
modification. From these results it can be observed that accuracy
has been improved for all TSP instances by several percent. More
importantly, the computational speed of the approach has been
increased significantly. A speedup of up to 2.8x is observed with
speedups increasing with the size of the TSP instance. Thus,
\emph{PartialACO} is demonstratively both faster and more
accurate.

\vspace{-0.5cm}

\begin{center}
\begin{table}[!h]
\footnotesize \centering \caption{Results from using standard
P-ACO approach with accuracy expressed as percentage difference
from known optimum. Results averaged over 100 random runs.}
\begin{tabular}{c
                S[table-number-alignment=center,separate-uncertainty,table-figures-uncertainty=1,table-figures-integer = 2, table-figures-decimal = 2]
                S[table-figures-integer = 1, table-figures-decimal = 2, table-column-width=15mm]
                S[table-figures-integer = 1, table-figures-decimal = 2]
                S[table-number-alignment=center,separate-uncertainty,table-figures-uncertainty=1, table-figures-integer = 3, table-figures-decimal = 2]
                }
\toprule
{\multirow{2}{1.5cm}{\centering TSP Instance}} & \multicolumn{3}{c}{Accuracy (\% Error)} & {\multirow{2}{2.5cm}{\centering Execution Time (in seconds)}}\\
\cmidrule{2-4} & {Average} & {Best} & {Worst} & \\
\midrule
pcb442 & 4.16 \pm 1.37 & 1.58 & 8.78 & 40.53 \pm 0.42 \\
d657 & 8.02 \pm 2.41 & 3.43 & 12.44 & 72.87 \pm 0.53 \\
rat783 & 4.08 \pm 1.13 & 2.27 & 7.82 & 96.23 \pm 0.81 \\
pr1002 & 7.88 \pm 1.49 & 5.18 & 12.47 & 145.59 \pm 0.91 \\
pr2392 & 13.47 \pm 1.13 & 10.91 & 15.98 & 688.13 \pm 2.86 \\
\bottomrule
\end{tabular} \centering
\label{tab:ACO}
\end{table}
\end{center}

\vspace{-2.0cm}

\begin{center}
\begin{table}[!h]
\footnotesize \centering \caption{Results from using
\emph{PartialACO} approach with accuracy expressed as percentage
difference from known optimum and relative speedup to using
standard P-ACO approach. Results averaged over 100 random runs.}
\begin{tabular}{c
                S[table-number-alignment=center,separate-uncertainty,table-figures-uncertainty=1,table-figures-integer = 2, table-figures-decimal = 2]
                S[table-figures-integer = 1, table-figures-decimal = 2, table-column-width=15mm]
                S[table-figures-integer = 1, table-figures-decimal = 2]
                S[table-number-alignment=center,separate-uncertainty,table-figures-uncertainty=1, table-figures-integer = 3, table-figures-decimal = 2]
                c
                }
\toprule
{\multirow{2}{1.5cm}{\centering TSP Instance}} & \multicolumn{3}{c}{Accuracy (\% Error)} & {\multirow{2}{2.5cm}{\centering Execution Time (in seconds)}} & {\multirow{2}{1.4cm}{\centering Relative Speedup}}\\
\cmidrule{2-4} & {Average} & {Best} & {Worst} &  & \\
\midrule
pcb442 & 2.72 \pm 0.67 & 1.14 & 4.75 & 17.94 \pm 0.26 & 2.26x \\
d657 & 4.33 \pm 0.73 & 2.88 & 7.17 & 31.68 \pm 0.31 & 2.30x \\
rat783 & 3.64 \pm 0.60 & 2.18 & 5.99 & 40.01 \pm 0.40 & 2.41x \\
pr1002 & 4.06 \pm 0.57 & 2.61 & 5.24 & 58.27 \pm 0.33 & 2.50x \\
pr2392 & 9.47 \pm 2.09 & 5.01 & 12.84 & 245.96 \pm 0.88 & 2.80x \\
\bottomrule
\end{tabular} \centering
\label{tab:cross}
\end{table}
\end{center}
\vspace{-0.5cm}

\begin{center}
\begin{table}[!h]
\footnotesize \centering \caption{Results from using
\emph{PartialACO} approach with restrictions on the maximum
permissable modification. Accuracy expressed as percentage
difference from known optimum with relative speedup to using
standard P-ACO approach reported. Results are averaged over 100
random runs.}
\begin{tabular}{c
                c
                S[table-number-alignment=center,separate-uncertainty,table-figures-uncertainty=1,table-figures-integer = 2, table-figures-decimal = 2]
                S[table-figures-integer = 1, table-figures-decimal = 2, table-column-width=15mm]
                S[table-figures-integer = 1, table-figures-decimal = 2]
                S[table-number-alignment=center,separate-uncertainty,table-figures-uncertainty=1, table-figures-integer = 3, table-figures-decimal = 2]
                c
                }
\toprule
{\multirow{2}{1.5cm}{\centering TSP Instance}}  & {\multirow{2}{2cm}{\centering Max. Modification}}  & \multicolumn{3}{c}{Accuracy (\% Error)} & {\multirow{2}{2.5cm}{\centering Execution Time (in seconds)}} & {\multirow{2}{1.4cm}{\centering Relative Speedup}}\\
\cmidrule{3-5} & & {Average} & {Best} & {Worst} &  & \\
\midrule
& 50\% & 4.90 \pm 1.31 & 2.31 & 9.48 & 9.01 \pm 0.22 & 4.50x \\
& 40\% & 6.79 \pm 1.57 & 3.64 & 10.11 & 7.44 \pm 0.18 & 5.45x \\
pcb442 & 30\% & 8.95 \pm 1.71 & 5.05 & 13.13 & 6.05 \pm 0.12 & 6.70x \\
& 20\% & 11.91 \pm 2.09 & 6.36 & 15.52 & 4.91 \pm 0.10 & 8.26x \\
& 10\% & 16.59 \pm 2.16 & 9.70 & 21.71 & 3.97 \pm 0.05 & 10.22x \\
\midrule
& 50\% & 7.29 \pm 1.00 & 5.14 & 9.66 & 14.48 \pm 0.22 & 5.03x \\
& 40\% & 8.40 \pm 1.34 & 5.34 & 11.30 & 11.62 \pm 0.15 & 6.27x \\
d657 & 30\% & 10.23 \pm 1.46 & 6.84 & 13.87 & 9.27 \pm 0.16 & 7.86x \\
& 20\% & 13.12 \pm 1.97 & 6.99 & 17.87 & 7.16 \pm 0.11 & 10.17x \\
& 10\% & 17.12 \pm 1.69 & 12.21 & 22.37 & 5.53 \pm 0.06 & 13.17x \\
\midrule
& 50\% & 7.67 \pm 1.38 & 3.52 & 11.03 & 17.66 \pm 0.16 & 5.45x \\
& 40\% & 9.62 \pm 1.20 & 4.73 & 12.41 & 14.08 \pm 0.15 & 6.83x \\
rat783 & 30\% & 11.34 \pm 1.43 & 7.46 & 15.95 & 10.91 \pm 0.18 & 8.82x \\
& 20\% & 13.46 \pm 1.68 & 9.45 & 16.77 & 8.27 \pm 0.13 & 11.64x \\
& 10\% & 17.09 \pm 1.53 & 11.78 & 20.97 & 5.98 \pm 0.08 & 16.10x \\
\midrule
& 50\% & 7.10 \pm 1.32 & 4.24 & 11.54 & 24.36 \pm 0.35 & 5.98x \\
& 40\% & 9.24 \pm 1.47 & 5.73 & 13.03 & 19.34 \pm 0.27 & 7.53x \\
pr1002 & 30\% & 10.52 \pm 1.70 & 6.12 & 13.41 & 14.70 \pm 0.19 & 9.91x \\
& 20\% & 12.75 \pm 1.84 & 7.96 & 16.98 & 10.93 \pm 0.16 & 13.33x \\
& 10\% & 16.32 \pm 1.57 & 11.43 & 20.44 & 7.83 \pm 0.12 & 18.61x \\
\midrule
& 50\% & 13.98 \pm 1.45 & 11.41 & 16.61 & 82.76 \pm 0.37 & 8.31x \\
& 40\% & 16.37 \pm 1.18 & 13.32 & 18.97 & 61.25 \pm 0.30 & 11.23x \\
pr2392 & 30\% & 18.01 \pm 1.54 & 14.15 & 21.14 & 43.45 \pm 0.17 & 15.84x \\
& 20\% & 20.01 \pm 1.56 & 15.33 & 22.22 & 29.84 \pm 0.17 & 23.06x \\
& 10\% & 21.92 \pm 1.02 & 19.87 & 24.56 & 20.74 \pm 0.18 & 33.18x \\
\bottomrule
\end{tabular} \centering
\label{tab:crossrange}
\end{table}
\end{center}

\vspace{-1cm}

Although the initial results from \emph{PartialACO} have
demonstrated a speed advantage with improved accuracy, it is
possible to increase the speed of the approach further. Currently,
a random part of the local best tour of an ant is preserved and
the rest exposed to ACO to modify it. However, the part that is
modified could be restricted to a maximum percentage of the
$l_{best}$ tour. For instance, a maximum percentage modification
of 50\% could be used thus for a 100 city problem at least part of
the $l_{best}$ tour consisting of 50 cities will be preserved.
Reducing the degree to which the $l_{best}$ tour of an ant can be
changed could also improve tour quality by increased tour
\emph{exploitation} whilst also increasing the speed advantage of
\emph{PartialACO}. Table \ref{tab:crossrange} demonstrates the
results from restricting the maximum amount that an ant's
$l_{best}$ tour can be modified whereby it can be observed that by
reducing the part of the $l_{best}$ tour that can be modified, the
average accuracy deteriorates with respect to the known optimum. A
potential reason for this is that the ants become trapped in local
optima, unable to improve their $l_{best}$ tour without a greater
degree of flexibility in tour construction. However, as expected,
reducing the degree to which a $l_{best}$ tour can be modified
increases the speed of the approach with up to a 33 fold increase
in speed observed when only allowing a maximum 10\% of $l_{best}$
tours to be modified at each iteration.

\vspace{-0.5cm}
\begin{center}
\begin{table}[!h]
\footnotesize \centering \caption{Results from using
\emph{PartialACO} approach with 0.95 probability and restrictions
on the maximum permissable modification. Accuracy expressed as
percentage difference from known optimum with relative speedup to
using standard P-ACO approach reported. Results are averaged over
100 random runs.}
\begin{tabular}{c
                c
                S[table-number-alignment=center,separate-uncertainty,table-figures-uncertainty=1,table-figures-integer = 2, table-figures-decimal = 2]
                S[table-figures-integer = 1, table-figures-decimal = 2, table-column-width=15mm]
                S[table-figures-integer = 1, table-figures-decimal = 2]
                S[table-number-alignment=center,separate-uncertainty,table-figures-uncertainty=1, table-figures-integer = 3, table-figures-decimal = 2]
                c
                }
\toprule
{\multirow{2}{1.5cm}{\centering TSP Instance}}  & {\multirow{2}{2cm}{\centering Max. Modification}}  & \multicolumn{3}{c}{Accuracy (\% Error)} & {\multirow{2}{2.5cm}{\centering Execution Time (in seconds)}} & {\multirow{2}{1.4cm}{\centering Relative Speedup}}\\
\cmidrule{3-5} & & {Average} & {Best} & {Worst} &  & \\
\midrule
& 50\% & 2.55 \pm 0.94 & 1.19 & 5.25 & 10.47 \pm 0.21 & 3.87x \\
& 40\% & 2.61 \pm 0.94 & 1.01 & 5.22 & 9.00 \pm 0.22 & 4.50x \\
pcb442 & 30\% & 3.26 \pm 1.41 & 1.26 & 7.15 & 7.63 \pm 0.16 & 5.31x \\
& 20\% & 3.35 \pm 1.34 & 1.43 & 7.95 & 6.37 \pm 0.13 & 6.36x \\
& 10\% & 3.96 \pm 1.67 & 1.55 & 8.98 & 5.28 \pm 0.11 & 7.67x \\
\midrule
& 50\% & 4.79 \pm 1.25 & 2.61 & 7.71 & 17.08 \pm 0.21 & 4.27x \\
& 40\% & 5.28 \pm 1.26 & 2.83 & 8.99 & 14.26 \pm 0.22 & 5.11x \\
d657 & 30\% & 5.65 \pm 1.33 & 2.95 & 9.03 & 12.05 \pm 0.20 & 6.05x \\
& 20\% & 6.34 \pm 1.39 & 3.19 & 9.30 & 9.95 \pm 0.17 & 7.32x \\
& 10\% & 7.27 \pm 1.64 & 3.38 & 11.39 & 8.15 \pm 0.13 & 8.94x \\
\midrule
& 50\% & 4.46 \pm 1.45 & 1.81 & 8.60 & 21.23 \pm 0.29 & 4.53x \\
& 40\% & 5.04 \pm 1.30 & 2.37 & 7.53 & 17.92 \pm 0.22 & 5.37x \\
rat783 & 30\% & 5.89 \pm 1.48 & 2.25 & 9.40 & 14.91 \pm 0.20 & 6.45x \\
& 20\% & 6.99 \pm 1.30 & 3.10 & 10.26 & 12.27 \pm 0.14 & 7.84x \\
& 10\% & 8.14 \pm 1.83 & 3.30 & 10.90 & 9.95 \pm 0.16 & 9.67x \\
\midrule
& 50\% & 4.75 \pm 1.33 & 2.28 & 8.43 & 30.05 \pm 0.30 & 4.84x \\
& 40\% & 5.28 \pm 1.16 & 2.91 & 7.97 & 24.96 \pm 0.28 & 5.83x \\
pr1002 & 30\% & 6.02 \pm 1.23 & 3.27 & 9.41 & 20.66 \pm 0.22 & 7.05x \\
& 20\% & 6.58 \pm 1.17 & 3.69 & 9.04 & 16.95 \pm 0.21 & 8.59x \\
& 10\% & 8.11 \pm 1.46 & 2.76 & 10.55 & 13.88 \pm 0.15 & 10.49x \\
\midrule
& 50\% & 10.58 \pm 1.07 & 8.31 & 12.54 & 111.98 \pm 0.49 & 6.15x \\
& 40\% & 10.86 \pm 1.11 & 7.69 & 13.10 & 90.58 \pm 0.36 & 7.60x \\
pr2392 & 30\% & 11.11 \pm 0.96 & 9.48 & 13.17 & 73.76 \pm 0.38 & 9.33x \\
& 20\% & 11.32 \pm 1.12 & 8.50 & 13.65 & 60.03 \pm 0.35 & 11.46x \\
& 10\% & 11.80 \pm 1.02 & 9.81 & 13.84 & 49.79 \pm 0.29 & 13.82x \\
\bottomrule
\end{tabular} \centering
\label{tab:crossprob}
\end{table}
\end{center}

\vspace{-1.0cm}

Clearly, restricting the maximum aspect of $l_{best}$ tours that
can be modified results in much faster speed but tour quality
suffers considerably as a result of being trapped in local optima.
A methodology is required to enable an ant to jump out of local
optima. In GAs, crossover is used with a given probability so
perhaps the same approach will benefit \emph{PartialACO}.
Therefore, it is proposed that an additional parameter is
introduced defining the probability that an ant will only
partially modify its $l_{best}$ tour. In the case an ant does not
partially modify its $l_{best}$ tour then it will construct a full
tour as standard P-ACO would. Table \ref{tab:crossprob} shows the
results from using a probability of an ant only partially
modifying its $l_{best}$ tour of 0.95. Comparing to Table
\ref{tab:crossrange}, improvements in the average tour accuracy to
the known optimum of each TSP instance are observed. However,
aside from the pcb442 problem, accuracy remains worse than the
results shown in Table \ref{tab:cross} with no restriction on the
degree by which $l_{best}$ tours can be modified. More
importantly, a significant reduction in the relative speedups is
observed even when using such a small probability of constructing
full tours.

\vspace{-0.2cm} \subsection{Incorporating Local Search} Given that
enabling ants to occasionally construct a full length tour to
break out of local optima has some beneficial effect, a better
alternative could be considered. Instead of using occasional full
length tour construction, a local search heuristic such as 2-opt
could be applied to tours with a given probability. Using 2-opt
will improve tours and by the swapping of edges between cities at
any point of the tour, ants could break out of local optima. To
test this theory, standard P-ACO is tested once again this time
using a probability of using 2-opt search of 0.001 with the other
parameters remaining the same. These results are shown in Table
\ref{tab:ACOLS} whereby significant improvements in accuracy are
observed over not using 2-opt. However, there is an increase in
execution time by as much as 33\% as 2-opt is a computationally
intensive algorithm of $O(n^{2})$ complexity. \vspace{-0.5cm}
\begin{center}
\begin{table}[!h]
\footnotesize \centering \caption{Results from using standard
P-ACO approach with a probability of 0.001 of using 2-opt local
search with accuracy expressed as percentage difference from known
optimum. Results are averaged over 100 random runs.}
\begin{tabular}{c
                S[table-number-alignment=center,separate-uncertainty,table-figures-uncertainty=1,table-figures-integer = 2, table-figures-decimal = 2]
                S[table-figures-integer = 1, table-figures-decimal = 2, table-column-width=15mm]
                S[table-figures-integer = 1, table-figures-decimal = 2]
                S[table-number-alignment=center,separate-uncertainty,table-figures-uncertainty=1, table-figures-integer = 3, table-figures-decimal = 2]
                }
\toprule
{\multirow{2}{1.5cm}{\centering TSP Instance}} & \multicolumn{3}{c}{Accuracy (\% Error)} & {\multirow{2}{2.5cm}{\centering Execution Time (in seconds)}}\\
\cmidrule{2-4} & {Average} & {Best} & {Worst} & \\
\midrule
pcb442 & 3.87 \pm 0.39 & 2.87 & 4.52 & 44.67 \pm 0.41 \\
d657 & 4.45 \pm 0.30 & 3.42 & 5.05 & 83.97 \pm 0.49 \\
rat783 & 5.20 \pm 0.29 & 4.24 & 5.83 & 110.43 \pm 0.63 \\
pr1002 & 5.56 \pm 0.32 & 4.65 & 6.18 & 170.48 \pm 0.95 \\
pr2392 & 7.47 \pm 0.27 & 6.50 & 7.90 & 834.08 \pm 5.71 \\
\bottomrule
\end{tabular} \centering
\label{tab:ACOLS}
\end{table}
\end{center}

\vspace{-1.9cm}

\begin{center}
\begin{table}[!h]
\footnotesize \centering \caption{Results from using
\emph{PartialACO} approach with a probability of 0.001 of using
2-opt local search. Results are averaged over 100 random runs.}
\begin{tabular}{c
                S[table-number-alignment=center,separate-uncertainty,table-figures-uncertainty=1,table-figures-integer = 2, table-figures-decimal = 2]
                S[table-figures-integer = 1, table-figures-decimal = 2, table-column-width=15mm]
                S[table-figures-integer = 1, table-figures-decimal = 2]
                S[table-number-alignment=center,separate-uncertainty,table-figures-uncertainty=1, table-figures-integer = 3, table-figures-decimal = 2]
                c
                }
\toprule
{\multirow{2}{1.5cm}{\centering TSP Instance}} & \multicolumn{3}{c}{Accuracy (\% Error)} & {\multirow{2}{2.5cm}{\centering Execution Time (in seconds)}} & {\multirow{2}{1.4cm}{\centering Relative Speedup}}\\
\cmidrule{2-4} & {Average} & {Best} & {Worst} &  & \\
\midrule
pcb442 & 1.64 \pm 0.30 & 0.74 & 2.32 & 21.26 \pm 0.31 & 2.10x \\
d657 & 2.32 \pm 0.31 & 1.20 & 3.10 & 39.65 \pm 0.40 & 2.12x \\
rat783 & 3.35 \pm 0.36 & 2.30 & 4.15 & 51.82 \pm 0.62 & 2.13x \\
pr1002 & 3.40 \pm 0.31 & 2.53 & 4.04 & 79.09 \pm 0.64 & 2.16x \\
pr2393 & 5.90 \pm 0.30 & 5.01 & 6.58 & 377.61 \pm 4.76 & 2.21x \\
\bottomrule
\end{tabular} \centering
\label{tab:crossLS}
\end{table}
\end{center}
\vspace{-1.45cm}
\begin{center}
\begin{table}[!h]
\footnotesize \centering \caption{Results from \emph{PartialACO}
approach using 0.001 probability of 2-opt and a range of maximum
modifications. Accuracy expressed as percentage difference from
known optimum with relative speedup. Results averaged over 100
random runs.}
\begin{tabular}{c
                c
                S[table-number-alignment=center,separate-uncertainty,table-figures-uncertainty=1,table-figures-integer = 2, table-figures-decimal = 2]
                S[table-figures-integer = 1, table-figures-decimal = 2, table-column-width=15mm]
                S[table-figures-integer = 1, table-figures-decimal = 2]
                S[table-number-alignment=center,separate-uncertainty,table-figures-uncertainty=1, table-figures-integer = 3, table-figures-decimal = 2]
                c
                }
\toprule
{\multirow{2}{1.5cm}{\centering TSP Instance}}  & {\multirow{2}{2cm}{\centering Max. Modification}}  & \multicolumn{3}{c}{Accuracy (\% Error)} & {\multirow{2}{2.5cm}{\centering Execution Time (in seconds)}} & {\multirow{2}{1.4cm}{\centering Relative Speedup}}\\
\cmidrule{3-5} & & {Average} & {Best} & {Worst} &  & \\
\midrule
& 50\% & 1.46 \pm 0.28 & 0.80 & 2.21 & 11.49 \pm 0.21 & 3.89x \\
& 40\% & 1.48 \pm 0.29 & 0.75 & 2.24 & 9.74 \pm 0.17 & 4.58x \\
pcb442 & 30\% & 1.49 \pm 0.33 & 0.76 & 2.42 & 8.15 \pm 0.19 & 5.48x \\
& 20\% & 1.47 \pm 0.32 & 0.65 & 2.25 & 6.63 \pm 0.11 & 6.74x \\
& 10\% & 1.80 \pm 0.37 & 0.87 & 2.86 & 5.12 \pm 0.09 & 8.73x \\
\midrule
& 50\% & 1.97 \pm 0.28 & 1.13 & 2.60 & 21.11 \pm 0.27 & 3.98x \\
& 40\% & 1.90 \pm 0.27 & 1.24 & 2.65 & 18.10 \pm 0.30 & 4.64x \\
d657 & 30\% & 1.80 \pm 0.28 & 1.22 & 2.39 & 15.15 \pm 0.29 & 5.54x \\
& 20\% & 1.71 \pm 0.28 & 0.94 & 2.38 & 12.28 \pm 0.22 & 6.84x \\
& 10\% & 1.86 \pm 0.26 & 0.86 & 2.41 & 9.32 \pm 0.22 & 9.01x \\
\midrule
& 50\% & 3.07 \pm 0.29 & 2.16 & 3.76 & 27.37 \pm 0.42 & 4.04x \\
& 40\% & 3.01 \pm 0.31 & 2.00 & 3.78 & 23.56 \pm 0.35 & 4.69x \\
rat783 & 30\% & 2.95 \pm 0.28 & 1.99 & 3.51 & 19.73 \pm 0.36 & 5.60x \\
& 20\% & 2.81 \pm 0.26 & 1.90 & 3.29 & 16.10 \pm 0.34 & 6.86x \\
& 10\% & 2.86 \pm 0.26 & 2.06 & 3.47 & 12.25 \pm 0.28 & 9.01x \\
\midrule
& 50\% & 2.93 \pm 0.31 & 1.77 & 3.44 & 42.47 \pm 0.59 & 4.01x \\
& 40\% & 2.81 \pm 0.32 & 2.06 & 3.54 & 36.40 \pm 0.55 & 4.68x \\
pr1002 & 30\% & 2.65 \pm 0.27 & 2.07 & 3.37 & 30.86 \pm 0.57 & 5.52x \\
& 20\% & 2.58 \pm 0.29 & 1.71 & 3.16 & 25.42 \pm 0.51 & 6.71x \\
& 10\% & 2.55 \pm 0.28 & 1.92 & 3.13 & 19.62 \pm 0.43 & 8.69x \\
\midrule
& 50\% & 5.46 \pm 0.27 & 4.86 & 5.99 & 204.69 \pm 3.67 & 4.07x \\
& 40\% & 5.38 \pm 0.28 & 4.73 & 6.06 & 179.68 \pm 3.79 & 4.64x \\
pr2392 & 30\% & 5.13 \pm 0.27 & 4.51 & 5.84 & 157.07 \pm 3.48 & 5.31x \\
& 20\% & 4.86 \pm 0.23 & 4.25 & 5.28 & 135.54 \pm 3.42 & 6.15x \\
& 10\% & 4.45 \pm 0.28 & 3.67 & 5.06 & 110.85 \pm 2.82 & 7.52x \\
\bottomrule
\end{tabular} \centering
\label{tab:crossrangeLS}
\end{table}
\end{center}

Table \ref{tab:crossLS} shows the results of the proposed
\emph{PartialACO} approach with the same probability of using
2-opt and no restriction to the portion of an ant's $l_{best}$
tour that can be modified. Improvements in accuracy are observed
for all the TSP instances over standard P-ACO. A speedup of a
little over two fold is also achieved, slightly less than that
when not using 2-opt. This is because a significant amount of
computational time is now spent within the 2-opt heuristic
reducing the advantage of \emph{PartialACO}.

Given the success of using 2-opt local search with
\emph{PartialACO}, the experiments restricting the degree to which
an ant can modify its $l_{best}$ tour can be repeated, the results
of which are shown in Table \ref{tab:crossrangeLS}. Now with 2-opt
local search, the accuracies from Table \ref{tab:crossLS} are all
improved upon by restricting the degree to which ants can modify
their $l_{best}$ tours. The point at which the best accuracy is
achieved favours smaller maximum modifications as the problem size
increases with only 10\% for the pr2392 problem although this
still enables partial tour modification of up to 239 cities. A
potential reason for the success of restrictive \emph{PartialACO}
when using 2-opt local search is that 2-opt derives high quality
tours whereby the subsequent iterations by ants are effectively
performing a localised search on these tours. The
\emph{exploitation} aspect of \emph{PartialACO} stems from
exploiting high quality tours which 2-opt local search assists but
it should be noted that the speedups are significantly reduced
when using 2-opt.

\vspace{-0.2cm}

\section{Applying \emph{PartialACO} to Larger TSP
Instances} \label{sec:LargeTSP}
\begin{figure}[!h]
\centering
\begin{tabular}{lrlr}
(a) & \includegraphics[width=0.3\textwidth]{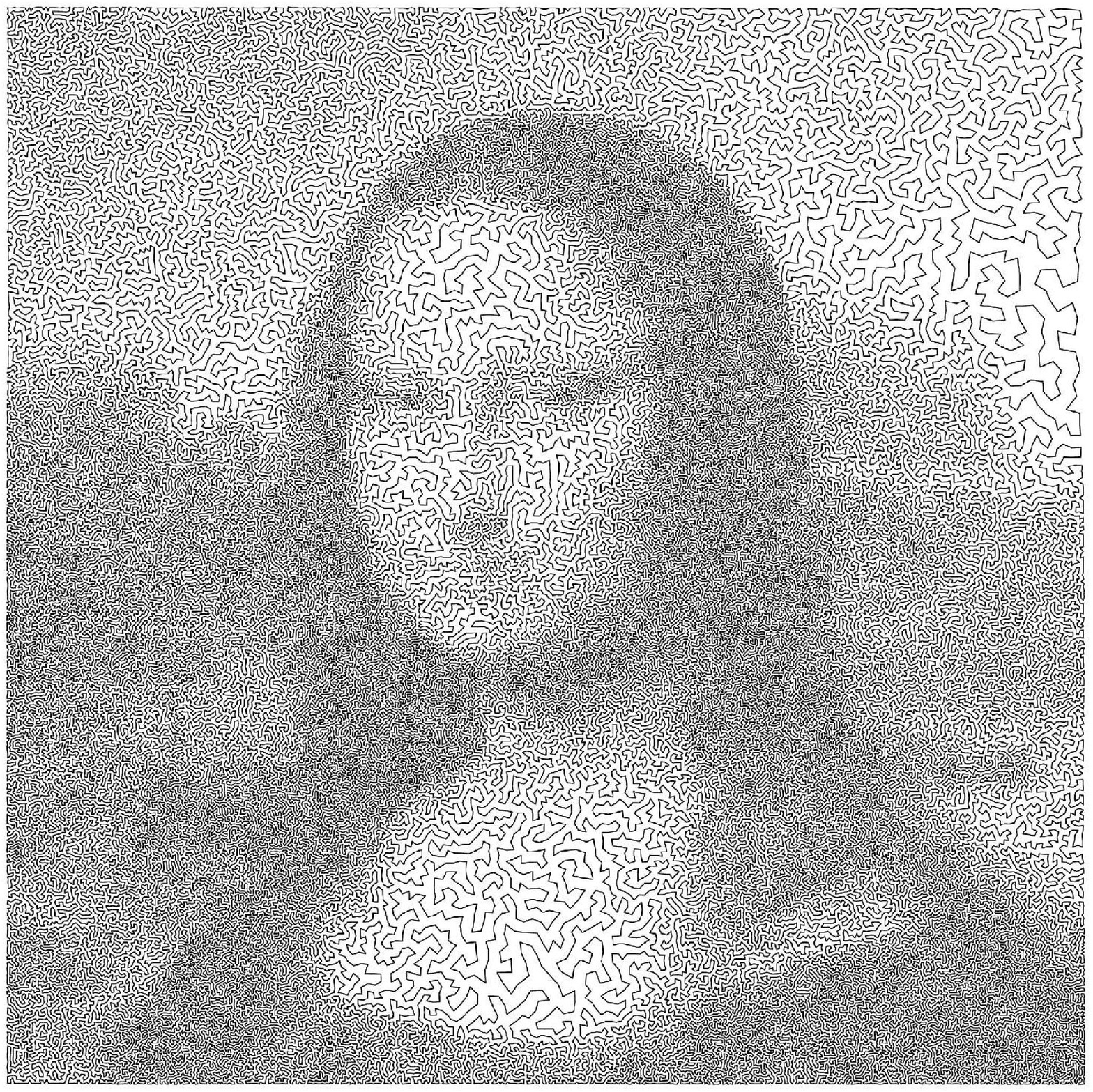} & \includegraphics[width=0.3\textwidth]{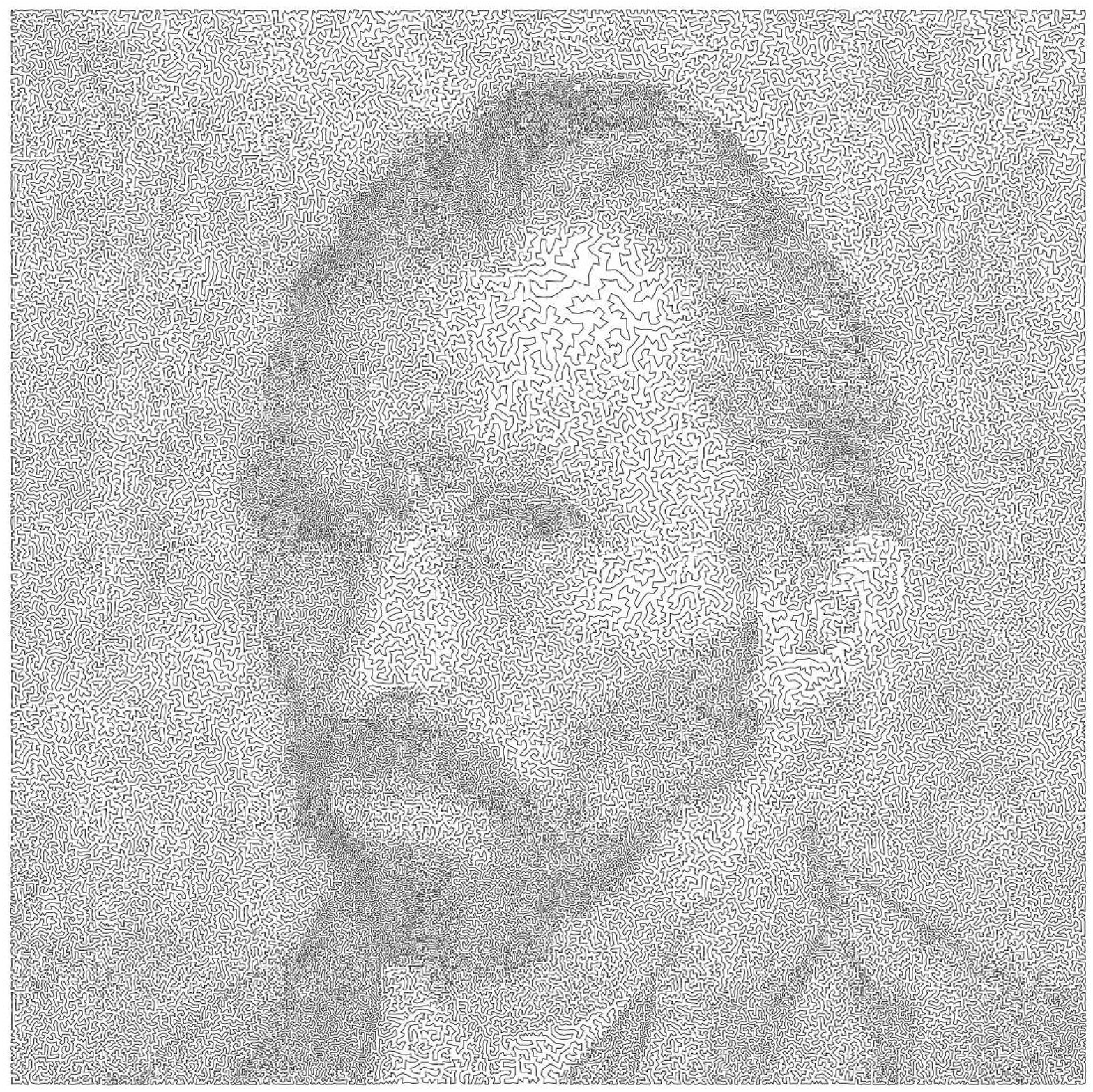} & (b)\\
(c) & \includegraphics[width=0.3\textwidth]{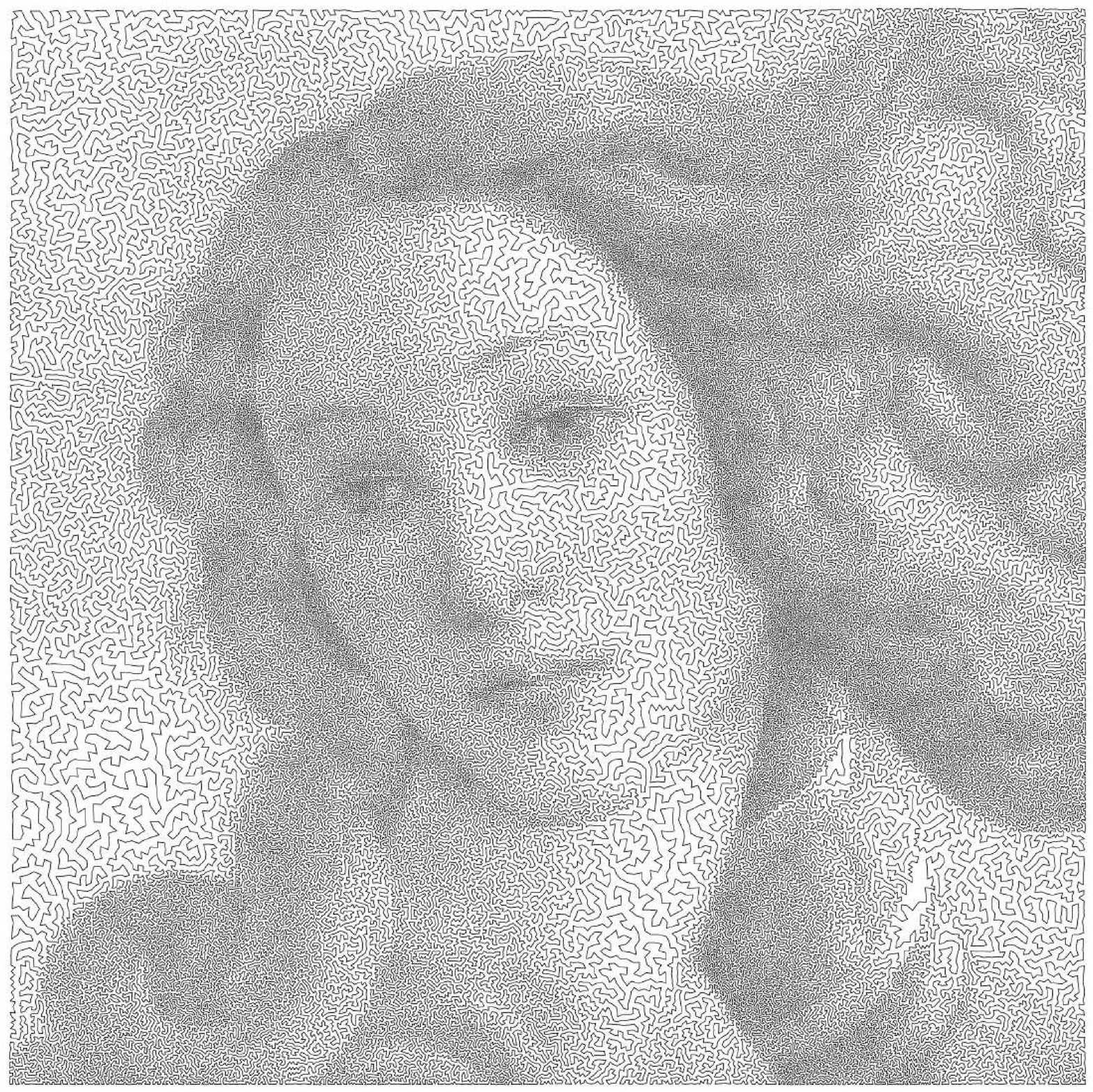} & \includegraphics[width=0.3\textwidth]{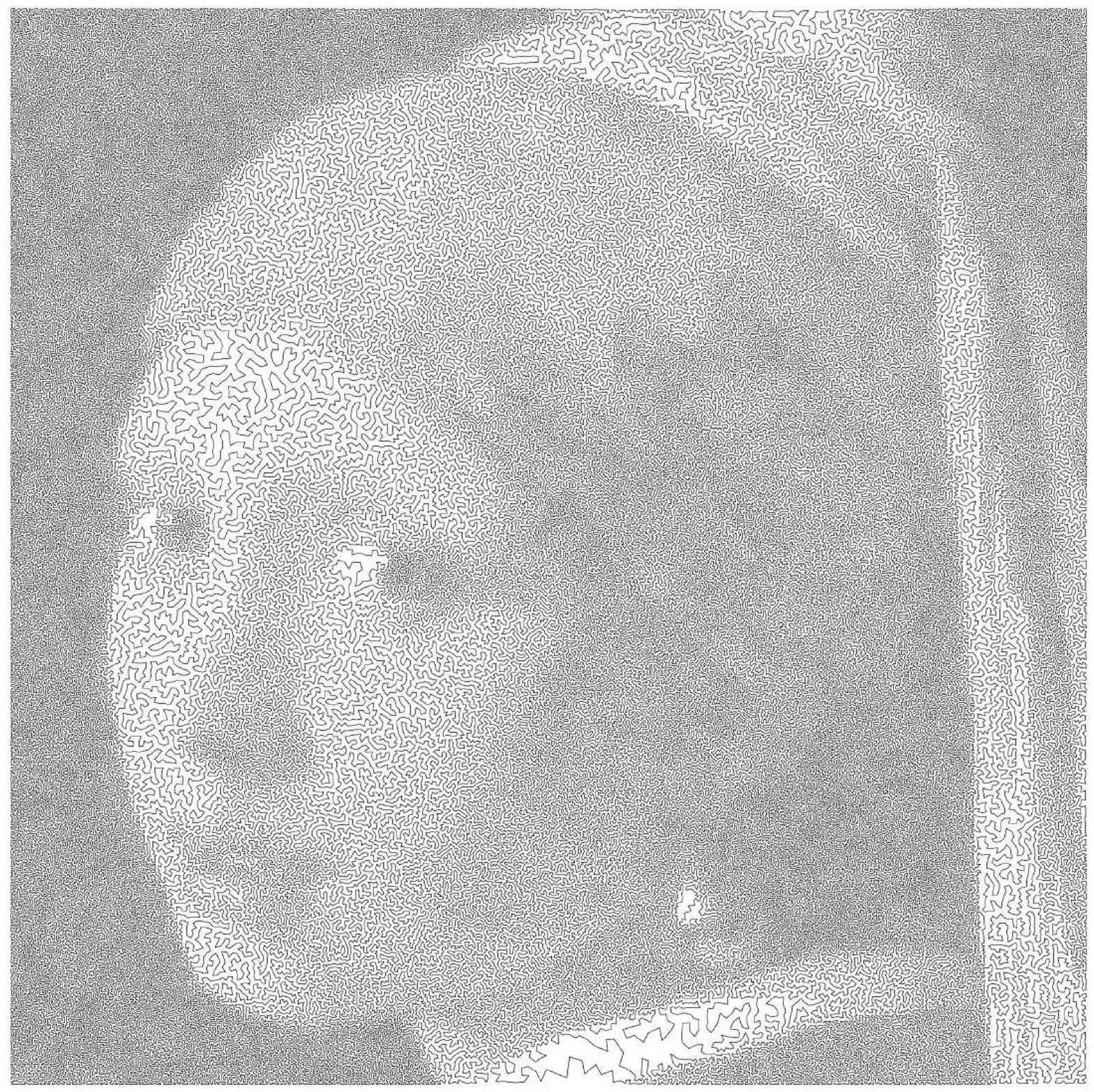} & (d)\\
\end{tabular}
\footnotesize \caption{Four classical art based TSP instances
(downloadable from http://www.math.uwaterloo.ca/tsp/data/art/),
(a) da Vinci's \emph{Mona Lisa} (100K cities), (b) Van Gogh's
\emph{Self Portrait 1889} (120K cities), (c) Botticelli's
\emph{The Birth of Venus} (140K cities) and (d) Vermeer's
\emph{Girl with a Pearl Earring} (200K cities)} \label{fig:Art}
\end{figure}

Now that the suitability of
\emph{PartialACO} has been demonstrated against regular size TSP
instances, it will now be tested against four much larger TSP
instances with hundreds of thousands of cities. These four large
TSP instances are based on famous works of art such as the
\emph{Mona Lisa} and the \emph{Girl with a Pearl Earring}. With
these TSP instances, the optimal tour when drawn in two dimensions
as a continual line will resemble the given famous art work (see
Figure \ref{fig:Art}).

\vspace{-0.5cm} \addtolength{\tabcolsep}{-1pt}
\addtolength{\tabcolsep}{1pt}

As previously, \emph{PartialACO} will be run for 100,00 iterations
with 16 ants and results averaged over 10 random runs. Moreover,
given the large scale of the problems under consideration, the
maximum aspect of the $l_{best}$ tour that can be modified by an
ant is now reduced to just 1\% of the number of cities which for
the Mona Lisa TSP is still 1,000 cities. 2-opt search will be
performed with a probability of 0.001. However, 2-opt is a
computationally expensive algorithm of $O(n^{2})$ complexity.
Consequently, 2-opt will be restricted to only considering
swapping edges that are within 500 cities of each other in the
current tour. This reduces the runtime of 2-opt significantly
although slightly reduces its effectiveness.

\begin{center}
\begin{table}[!h]
\footnotesize \centering \caption{Results from testing
\emph{PartialACO} on four art TSP instances. Accuracy expressed as
percentage difference from best known tour. Results averaged over
10 runs.}
\begin{tabular}{c
                S[table-number-alignment=center,separate-uncertainty,table-figures-uncertainty=1,table-figures-integer = 2, table-figures-decimal = 2]
                S[table-figures-integer = 1, table-figures-decimal = 2, table-column-width=15mm]
                S[table-figures-integer = 1, table-figures-decimal = 2]
                S[table-number-alignment=center,separate-uncertainty,table-figures-uncertainty=1, table-figures-integer = 1, table-figures-decimal = 2]
                }
\toprule
{\multirow{2}{2.5cm}{\centering TSP Instance}} & \multicolumn{3}{c}{Accuracy (\% Error)} & {\multirow{2}{2.5cm}{\centering Execution Time (in hours)}}\\
\cmidrule{2-4} & {Average} & {Best} & {Worst} & \\
\midrule
mona-lisa100K & 5.45 \pm 0.07 & 5.36 & 5.58 & 1.07 \pm 0.02 \\
vangogh120K & 5.82 \pm 0.10 & 5.70 & 6.01 & 1.45 \pm 0.03 \\
venus140K & 5.81 \pm 0.14 & 5.60 & 6.05 & 2.09 \pm 0.06 \\
earring200K & 7.20 \pm 0.18 & 6.91 & 7.39 & 5.06 \pm 0.14 \\
\bottomrule
\end{tabular} \centering
\label{tab:artresults}
\end{table}
\end{center}

\vspace{-1.5cm}

\begin{center}
\begin{table}[!h]
\footnotesize \centering \caption{Results from testing standard
P-ACO against the four large art based TSP instances for the
timings reported in Table \ref{tab:artresults}. Accuracy is
expressed as the percentage difference from the best known
solution. The number of iterations is shown enabling a speedup of
\emph{PartialACO} to be ascertained. Results averaged over 10
runs.}
\begin{tabular}{c
                S[table-number-alignment=center,separate-uncertainty,table-figures-uncertainty=1,table-figures-integer = 3, table-figures-decimal = 2]
                S[table-figures-integer = 4, table-figures-decimal = 2, table-column-width=15mm]
                S[table-figures-integer = 4, table-figures-decimal = 2]
                S[table-number-alignment=center,separate-uncertainty,table-figures-uncertainty=1, table-figures-integer = 4, table-figures-decimal = 2]
                c
                }
\toprule
{\multirow{2}{2.5cm}{\centering TSP Instance}}  & \multicolumn{3}{c}{Accuracy (\% Error)} & {\multirow{2}{1.4cm}{\centering Average Iterations}} & {\multirow{2}{2.5cm}{\centering Relative Speedup by \emph{PartialACO}}}\\
\cmidrule{2-4} & {Average} & {Best} & {Worst} & & \\
\midrule
mona-lisa100K & 13.50 \pm 0.49 & 12.62 & 14.46 & 248.10 \pm 4.41 & 403.06x \\
vangogh120k & 14.04 \pm 0.54 & 13.25 & 15.07 & 157.00 \pm 1.76 & 636.94x \\
venus140k & 14.48 \pm 2.24 & 13.04 & 20.71 & 105.30 \pm 3.77 & 949.67x \\
earring200k & 16.71 \pm 3.25 & 13.97 & 21.46 & 83.40 \pm 0.52 & 1199.04x \\
\bottomrule
\end{tabular} \centering
\label{tab:artresultsBasicACO}
\end{table}
\end{center}
\vspace{-0.75cm}

The results of executing the \emph{PartialACO} technique on the
large scale TSP instances are shown in Table \ref{tab:artresults}
whereby it can be observed that tours with an average error
ranging between 5-7\% of the best known optima are found.
Regarding runtime, \emph{PartialACO} finds these tours with just a
few hours of computational time using a single multi-core CPU. To
clearly demonstrate the effectiveness of \emph{PartialACO} a
comparison is made with the standard P-ACO approach. Given the
likely increase in runtime it would not be feasible to execute for
100,000 iterations. Consequently, a time limited approach is used
whereby the standard P-ACO approach is run for the same degree of
time as the results from Table \ref{tab:artresults} and the number
of iterations achieved recorded which provides a relative speedup
achieved by \emph{PartialACO}. These results are shown in Table
\ref{tab:artresultsBasicACO} whereby it can be observed that much
worse accuracy is achieved which is to be expected as P-ACO only
ran for a few hundred iterations. Furthermore, a speedup of up to
1200x is demonstrated by \emph{PartialACO}. To understand this
speedup the number of required edge comparisons to unvisited
cities to determine which city to visit next by an ant building a
tour must be considered. The number of comparisons an ant will
make relates to a triangle progression sequence defined by
$(n(n+1))/2$. Thus, an ant building a complete tour for a 100k
city problem will perform approximately $5\times10^{9}$ edge
comparisons. However, if only 1\% of a tour is modified an ant
will only perform $5\times10^{5}$ comparisons, a 10,000 fold
efficiency saving. However, not all of this saving will be
realised as a result of computational factors such as the use of
2-opt. The 10,000 fold efficiency only applies to the \emph{tour
construction} aspect of ACO hence the lower reported speedups.

\begin{figure}[!h]
\centering
\begin{tabular}{lr}
\includegraphics[width=0.45\textwidth]{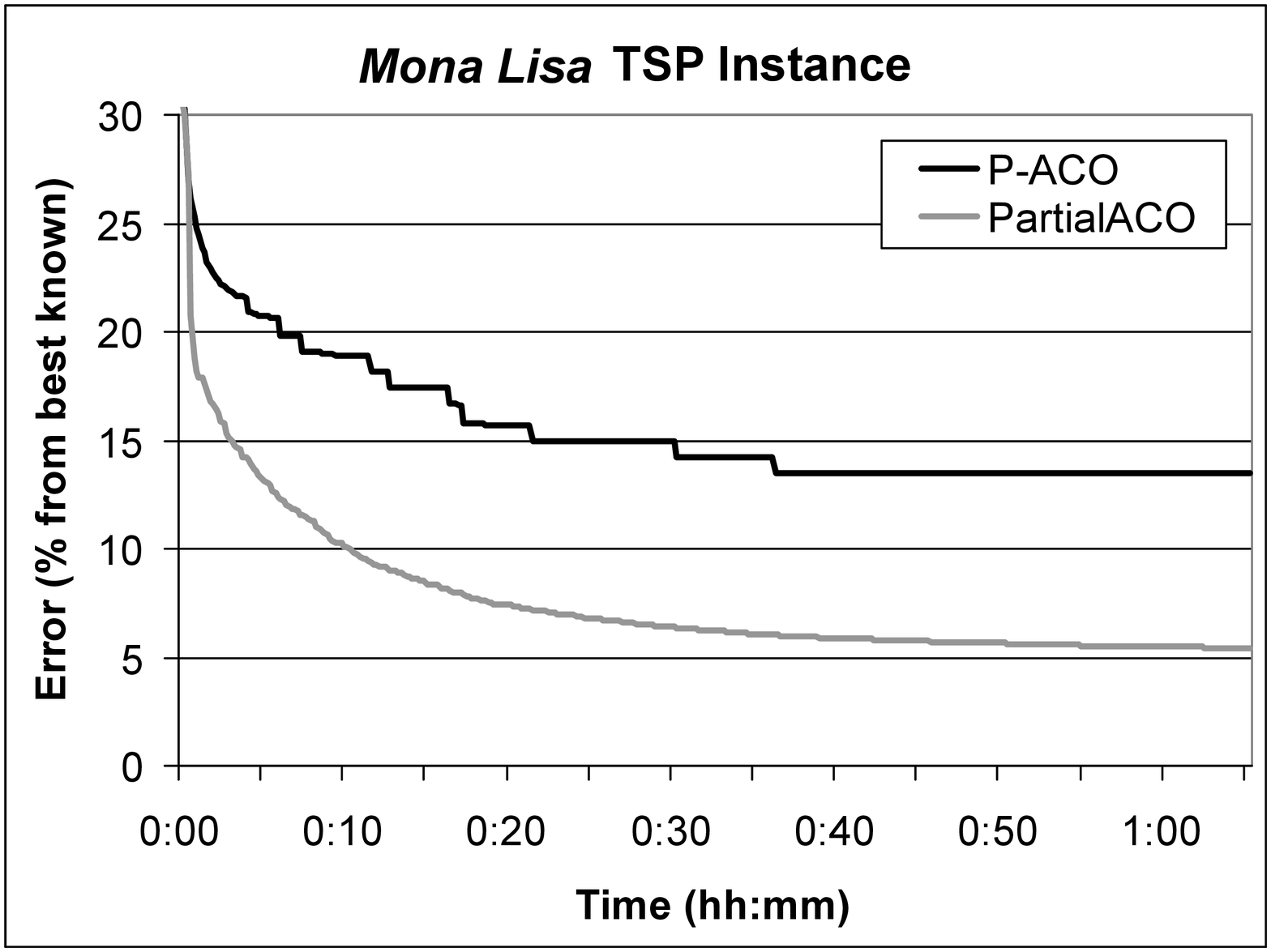} & \includegraphics[width=0.45\textwidth]{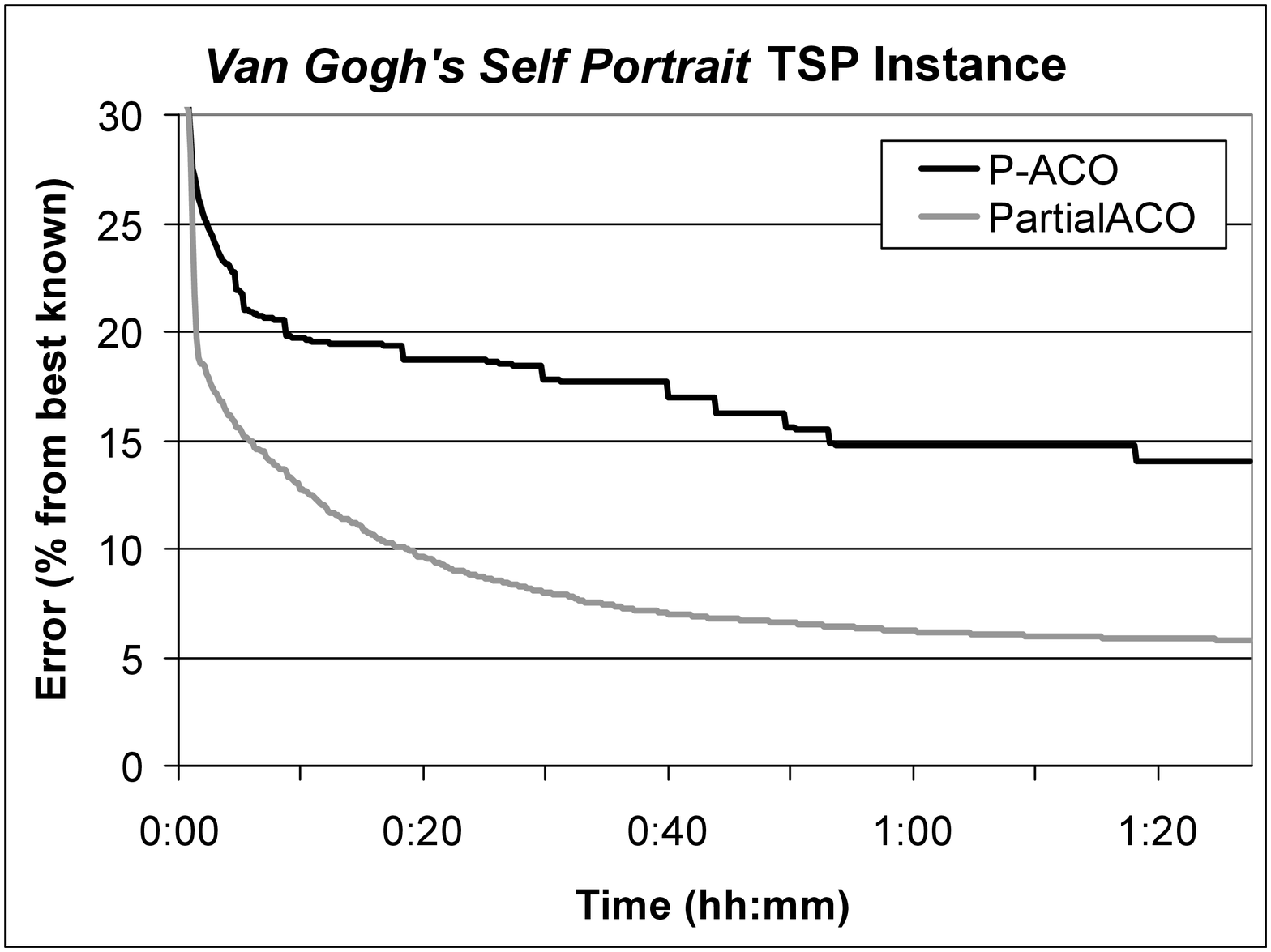}\\
\includegraphics[width=0.45\textwidth]{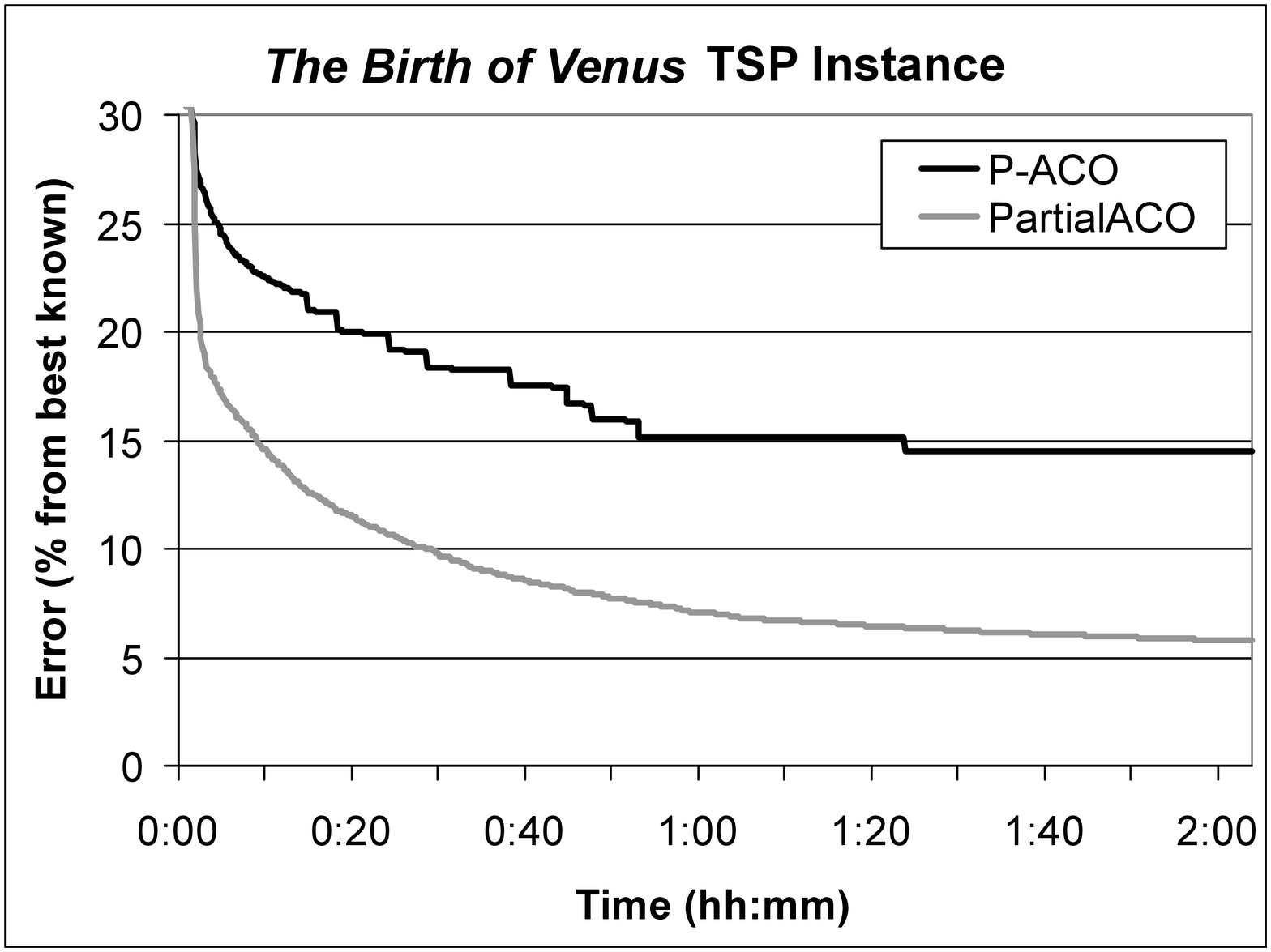} & \includegraphics[width=0.45\textwidth]{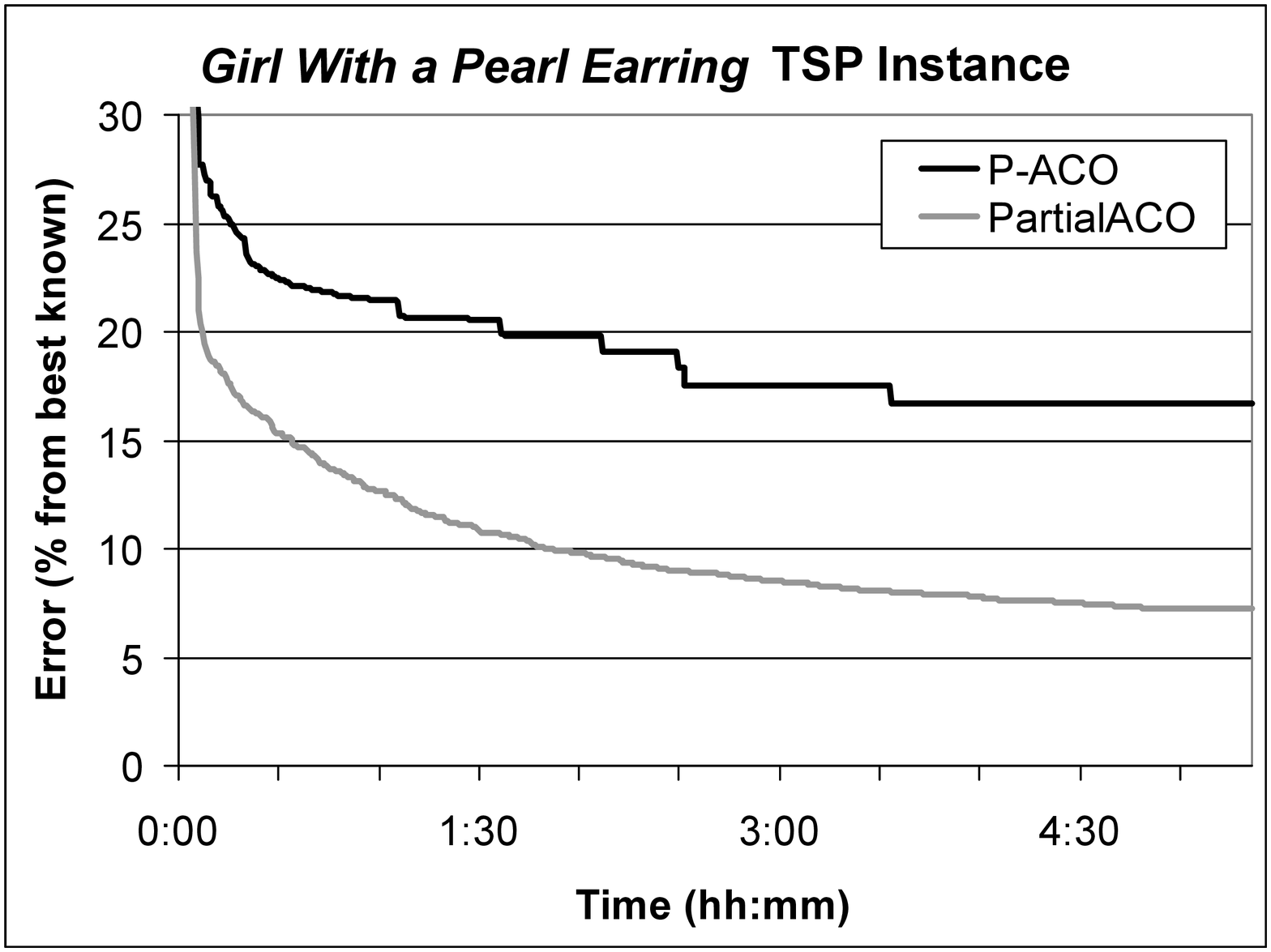}\\
\end{tabular}
\footnotesize \caption{The average convergence rates over time for
the P-ACO and \emph{PartialACO} techniques and each of the art
based TSP instances} \label{fig:Convergence}
\end{figure}
\vspace{-0.5cm}

Further evidence of the effectiveness of \emph{PartialACO} is
demonstrated by considering the convergence rates over time for
each of the art based TSP instances as shown in Figure
\ref{fig:Convergence}. \emph{PartialACO} clearly converges much
faster than the P-ACO approach for all four problems. Indeed,
inspection of the largest problem instance, \emph{The Girl With
the Pearl Earring}, shows that the \emph{PartialACO} technique
achieves the same accuracy in minutes that P-ACO takes hours to
achieve. This convergence speed is simply as a result of
\emph{PartialACO} being able to perform many more iterations of
the ant tours, indeed thousands, in a short space of time. In
fact, it can be argued that in terms of iterations
\emph{PartialACO} converges slower but given that time is a more
important factor, \emph{PartialACO} is the better approach.

\section{Related Work}
It is acknowledged that ACO is computationally intensive. Indeed,
even the original author of ACO was aware of the computational
complexity proposing a variant known as Ant Colony System (ACS)
\cite{dorigo:1997} whereby the neighbourhood of unvisited cities
is restricted. A \emph{candidate list} approach is used whereby at
each decision point made by an ant, only the closest cities are
considered. If these have already been visited then normal ACO
used. This approach significantly reduces the computational
complexity. ACS is also similar to \emph{PartialACO} in that with
a high probability an ant takes the edge with the greatest level
of combined pheromone and heuristic information improving the
speed. However, ACS still requires a full pheromone matrix and to
needs to search for the edge with the greatest level to choose the
next city to visit.

The main area of research into speeding up ACO though has been
through parallel implementations. ACO is naturally parallel such
that ants can construct tours simultaneously. Early works such as
Bullnheimer et al. \cite{bullnheimer:1998}, Delisle et al.
\cite{delisle:2001} and Randall and Lewis \cite{randall:2002}
relied on distributing ants to processors using a master-slave
methodology. In recent years the focus on speeding up ACO has been
on utilising Graphical Processor Units (GPUs) consisting of
thousands of SIMD processors. Bai et al. were the first to
implement MAX-MIN ACO for the TSP on a GPU achieving a 2.3x
speedup \cite{bai:2009}. More notable works include Del{\'e}Vacq
et al. who compare parallelisation strategies for MAX-MIN ACO on
GPUs \cite{delevacq:2013}, Cecelia et al. who present an
\emph{Independent Roulette} approach to better exploit data
parallelism for ACO on GPUs \cite{cecilia:2013} and Dawson and
Stewart who introduce a \emph{double spin} ant decision
methodology when using GPUs \cite{dawson:2013}. However, ACO is
not ideally suited to GPUs and these papers can only report
speedups ranging between 40-80x over a sequential implementation.

\section{Conclusions}
This paper has addressed the issues associated with applying ACO
to large scale TSP instances, namely reducing memory constraints
and substantially increasing execution speed. A new variant of ACO
was introduced, \emph{PartialACO}, based upon P-ACO which
dispenses with the pheromone matrix, the memory overhead.
Moreover, \emph{PartialACO} only partially modifies the best tour
found by each ant akin to crossover in GAs. \emph{PartialACO} was
demonstrated to significantly improve the computational speed of
ACO and the accuracy by reducing the computational complexity and
the probabilistic chance of ants making poor choices of cities to
visit. Consequently, \emph{PartialACO} was applied to large scale
TSP instances of up to 200K cities achieving accuracy of 5-7\% of
the best known tours with a speed of up to 1200 times faster than
that of standard P-ACO. \emph{PartialACO} is a first step to
deploying ACO on large scale TSP instances and further work is
required to improve its accuracy to compete with a GA approach
\cite{honda:2013} although it should be noted that this work uses
a supercomputer. Further analysis of the parameters balancing
speed vs. accuracy could help to improve the technique such as
reducing the maximum permissable modification of tours for speed
and increasing the iterations. Moreover, a dynamic approach may be
best whereby initially only small modifications are allowed but as
time progresses the permissible modification increases to avoid
being trapped in local optima.

\section{Acknowledgement}
This is a pre-print of a contribution published in Chao F.,
Schockaert S., Zhang Q. (eds) Advances in Computational
Intelligence Systems. UKCI 2017, Advances in Intelligent Systems
and Computing, vol. 650 published by Springer. The definitive
authenticated version is available online via
https://doi.org/10.1007/978-3-319-66939-7\_9.

\scriptsize
\bibliographystyle{splncs03}
\bibliography{ApplyingACOToLargeScaleTSPInstances}
\end{document}